\documentclass[letterpaper, 10 pt, conference]{ieeeconf}  
\IEEEoverridecommandlockouts                              
\usepackage{graphics,graphicx}           
\usepackage{amsmath,amssymb}    
\usepackage{newtxtext,newtxmath}
\usepackage{bm}
\usepackage{booktabs}
\usepackage{color,xspace}
\usepackage{subfigure}
\usepackage{balance}
\usepackage{cite}
\usepackage{color,soul}
\usepackage{mathtools}
\usepackage{makeidx}
\usepackage{robustindex}
\usepackage{float}
\usepackage{multirow}
\usepackage{textcomp}
\usepackage{url}
\makeindex

\title{\LARGE \bf
Flight Recovery of MAVs with Compromised IMU
}

\author{Zhan Tu\textsuperscript{+}, 
Fan Fei\textsuperscript{+}, 
Matthew Eagon\textsuperscript{+}, 
Dongyan Xu\textsuperscript{++}, 
and Xinyan Deng\textsuperscript{+}
\thanks{\textsuperscript{+}School of Mechanical Engineering, Purdue University.}
\thanks{\textsuperscript{++}Department of Computer Science, Purdue University.}
}

\begin{document}

\maketitle
\thispagestyle{empty}
\pagestyle{empty}

\begin{abstract} 
Micro Aerial Vehicles (MAVs) rely on onboard attitude and position sensors for autonomous flight. Due to their size, weight, and power (SWaP) constraints, most modern MAVs use miniaturized inertial measurement units (IMUs) to provide attitude feedback, which is critical for flight stabilization and control. However, recent adversarial attack studies have demonstrated that many commonly used IMUs are vulnerable to attacks exploiting their physical characteristics. Conventional redundancy-based approaches are not effective against such attacks because redundant IMUs have the same or similar physical vulnerabilities. In this paper, we present a novel fault-tolerant solution for IMU compromised scenarios, using separate position and heading information to restore the failed attitude states. Rather than adding more IMU alternatives for recovery, the proposed method is intended to minimize any modifications to the existing system and control program. Thus, it is particularly useful for vehicles that have tight SWaP constraints while requiring simultaneous high performance and safety demands. To execute the recovery logic properly, a robust estimator was designed for fine-grained detection and isolation of the faulty sensors. The effectiveness of the proposed approach was validated on a quadcopter MAV through both simulation and experimental flight tests.
\end{abstract}
\section{Introduction} \label{sec:intro}
In recent years, autonomous Micro Aerial Vehicles (MAVs) equipped with onboard sensors have drawn great interest and wide usage in many applications such as search and rescue, surveying and inspection, precision agriculture, aerial photography, and hobbies\cite{tomic2012toward,hodgson2016precision,beloev2016review,ozaslan2017autonomous}. To achieve 6 degrees of freedom (3-axis position and 3-axis attitude) controlled flight, a typical sensory system of MAVs consists of attitude and position sensors for essential feedback to the flight controller \cite{mahony2012multirotor, zhang2014loam}. Thanks to Micro-Electro-Mechanical-Systems (MEMS) technology, most modern MAVs use MEMS-based inertial measurement units (IMUs) for attitude sensing, which accommodates the size, weight, and power (SWaP) constraints in such vehicle design and manufacturing \cite{mahony2012multirotor}. IMUs can support more than just attitude sensing and vehicle control. The many parameter measurements of these IMUs, including 3-axis vehicle body accelerations, 3-axis body angular velocities, and sometimes 3-axis magnetic field, have also been used to compensate for errors or failures in position sensing to enhance system robustness \cite{levi1996dead,svacha2018inertial}. From a flight safety and system security standpoint, such heavy reliance on the IMU provides a clear target for adversarial attacks \cite{kim2012cyber, javaid2012cyber,son2015rocking,trippel2017walnut}. In fact, adversarial attack methodologies targeting MAV IMUs are evolving rapidly.

To date, attacks targeted at IMUs are not limited to traditional hacking in the cyber domain \cite{kim2012cyber,javaid2012cyber}, but are progressively utilizing sensors' physical vulnerabilities to construct more effective and stealthy attacks \cite{son2015rocking, trippel2017walnut}. As presented in \cite{son2015rocking}, Son et al. have successfully disrupted the operation of drones using acoustic noise of tailored frequencies to trigger IMUs to resonate, causing sensor failure. Trippel et al. have manipulated the IMU readings with a similar approach and demonstrated the potential to take complete control of the vehicle and misguide it \cite{trippel2017walnut}. These state-of-the-art attacks cause sensor faults and provide a challenge for the two major aspects of current sensor fault-tolerant design: fault detection and isolation (FDI), and fault recovery (FR). 

Regarding FDI, the interest of past research has been to detect anomalies in vehicle states \cite{blanke2006diagnosis,qi2013fault,choi2018detecting}, which is not generally equivalent to specifying the physical level sensor failure. For MAVs, different sensor faults/attacks may induce similar flight state irregularities, e.g., attitude twitching can originate from either a compromised IMU or a flawed position sensor. This ambiguity complicates pinpointing the failure source and may mislead previous state-based FDI methods to carry out incorrect isolation in some sensor-failure scenarios of interest. Recently, the in-depth studies have considered FDI of sensor hardware by diagnosing the drift and abnormal noise of the sensor readings \cite{qi2007application,avram2015imu,avram2017quadrotor}. These signal processing-based approaches typically require a certain model of the sensor measurements, such as correlation function or frequency spectrum, to aid in estimation and fault detection. It is difficult for such methods to cover arbitrary flight conditions and scenarios. 

For FR, redundancy was previously considered an adequate strategy to handle sensor spoofing. However, redundant sensors are not sufficient when the vehicle encounters some attacks against the general physical vulnerabilities of MEMS IMUs since redundant inertial sensors would be subject to the same detrimental effects of resonating from external excitation. Separate IMUs could each fail with an acoustic attack exploiting their resonant frequencies. Adding different types of backup sensors with significantly different physical characteristics could be another potential solution to this problem \cite{chen2008design,achtelik2011onboard,faessler2015automatic,gowda2016tracking}. However, considering MAVs' SWaP constraints and the usage restrictions of the added sensors, such as the light condition for visual sensors, implementing these alternatives may still show limited performance.

To address these issues, we propose a fault-tolerant solution for MAVs equipped with inertial and position sensors with no redundancy. In the case of inertial sensor failure, the proposed solution uses the remaining position and heading information to restore the compromised attitude states. This method is designed as a two-step scheme, i.e., first running a fine-grained IMU FDI design and isolating the compromised sensors, then triggering a FR logic to function without IMU feedback accordingly. With such a design, the system robustness can be enhanced without re-designing the control program and hardware. Particularly, in the FDI module, we adopt a smooth variable structure filter (SVSF)\cite{habibi2007smooth} for state estimation. More critically, in the estimation process, we utilize the chattering signal from the estimated physical parameters (mass and inertia) of the vehicle to pinpoint the IMU faults/attacks when the traditional model-based FDI is invalid \cite{son2015rocking,blanke2006diagnosis,qi2013fault,avram2015imu,avram2017quadrotor,choi2018detecting}. This method goes deep into the sensor hardware fault detection without requiring additional modeling work. For FR, we present a geometrical approach that makes full use of the remaining sensors' information to reconstruct the lost attitude states. This method is particularly useful for vehicles with stringent SWaP constraints and has the potential to be used in conjunction with other fault-tolerant methods. With our approach, whether the fault came about from malicious attack or some other source, the functionality of the FDI and FR are not affected, namely, `fault' can generally be interchanged with malicious `attack' in many cases. 

In this work, we first introduce the physical test platform in section.\ref{sec:platform}. The sensor FDI and FR methods are then described in detail in section.\ref{sec:sensor FDI} and section.\ref{sec:FR}, respectively. Finally, in section.\ref{sec:result}, the effectiveness of the proposed approach is validated through both simulation and experimental flight tests using a quadcopter MAV as shown.

\section{Test Platform Description} \label{sec:platform}

\begin{figure}[t]
\begin{center}
\includegraphics[trim = 0mm 0mm 0mm 0mm, clip,width=0.8\columnwidth]{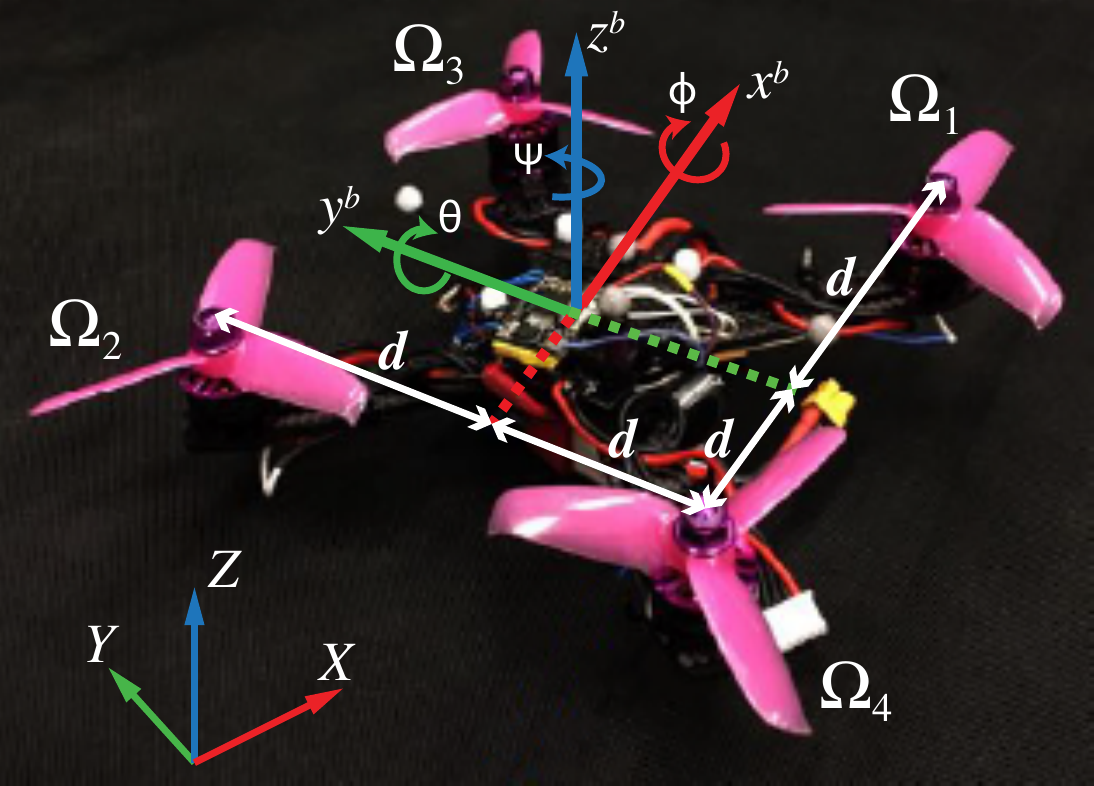}
\caption{Test platform coordinate definition. In this work, a small-sized racing drone is chosen as the test platform. $d$ denotes the distances from the motor to the center of mass.}
\label{fig:platform}
\end{center}
\end{figure}

As shown in Fig.\ref{fig:platform}, we use a quadrotor MAV as our test platform. The vehicle weighs 125 grams and has a 100mm wheelbase. The onboard system incorporates four motor drivers, an STM32 microcontroller with ARM-CortexM4 core, an MPU-6050 IMU that consists of a 3-axis gyroscope and accelerometer, and an extended compass HMC5883L for heading direction sensing. Position feedback is provided by an external motion capture system: VICON system (www.vicon.com). The entire setup constructs a typical MAV sensory system, wherein the position and attitude feedback come from different sensors, with different communication channels, and different signal frequencies corresponding to the relevant system bandwidth. These discrepancies are detailed in Table.\ref{tab:Att vs Pos}. The roll, pitch, and yaw are of the vehicle are outputs derived using values from all of the sensors in the IMU. Such sensor fusion lowers the uncertainty of the measured values of these parameters.

\begin{table}[h]
\footnotesize
\caption{Details of onboard Sensors.} \label{tab:Att vs Pos}
\setlength{\tabcolsep}{1mm}{
\begin{tabular}{lccc} 
\toprule   
Feedback 	& Sensor	& Communication 	& Frequencies\\   
\midrule  
$\ddot{x}^b,\ddot{y}^b,\ddot{z}^b$&IMU (Accelerometer) &I2C &1000Hz\\ 
$p,q,r$  & IMU (Gyroscope) & I2C & 1000Hz\\ 
$\phi,\theta$  & IMU (Sensor fusion) & I2C & 400Hz\\  
$\psi$ 	& Compass+IMU (Sensor fusion) & SPI	& 100Hz\\ 
$x,y,z$  & VICON  & Serial & 50Hz\\ 
\bottomrule  
\end{tabular}}
\end{table}


The coordinates definition of the vehicle and inertial frame is shown in Fig.\ref{fig:platform}. The vehicle dynamics is modeled as 

\begin{equation} \label{eq:body_dynamics}
\begin{aligned}
&\dot{\bm{P}}=\bm{V},& &m\ddot{\bm{P}}=\bm{R}\bm{f}^b+m \bm{g},\\
&\dot{\bm{R}} = \bm{R} [ \bm{\omega}_{\times}^b ],& &\bm{J} \dot{\bm{\omega}}^b = \bm{\tau}^b - \bm{\omega}^b \times \bm{J} \bm{\omega}^b. \\
\end{aligned}
\end{equation}
where $\bm{P}=[x,y,z]^T$ is the position vector of the vehicle in the inertial frame $XYZ$ which is defined by North-East-Up (NEU); 
$\bm{V}=[\dot{x}, \dot{y}, \dot{z}]^T$ is the velocity vector of the vehicle; 
$m$ is the total mass of the vehicle; 
$\bm{g}=[0,0, -\text{g}]^T$ is the gravitational acceleration vector; 
$\bm{R}$ is the rotation matrix; 
$\bm{J}$ is the inertia matrix of the vehicle;
define a generic symbol '$\bullet$'; $[\bullet_{\times}]$ denotes the skew-symmetric matrix mapping from vector dot product to vector cross product;
$\bm{\bullet}^b$ represents the vector in the body frame $x^by^bz^b$, e.g., the thrust vector $\bm{f}^b=[0,0,f_z]^T$, the vehicle angular velocity $\bm{\omega}^b = [p,q,r]^T$, and the 3-axis torque vector $\bm{\tau}^b=[\tau_x, \tau_y, \tau_z]^T$. 

\section{Faulty Sensor Detection and Isolation} \label{sec:sensor FDI}

Accurately pinpointing the flawed sensors is critical to execute the fault recovery logic correctly and ensure flight safety. In this section, we present a novel design for fine-grained sensor FDI. By monitoring the variations of the physical parameters of the vehicle, e.g., mass and 3-axis moment of inertia, the proposed method can explicitly identify the faulty sensors in some scenarios that the traditional model-based methods cannot. The proposed FDI approach is validated as shown in section.\ref{sec:result}.

\subsection{State Estimation and Residual Calculation} \label{subsec: Basic SVSF}

We first implemented an SVSF as a baseline estimator to allow state residual checking, using it as a conventional approach of state anomalies detection \cite{blanke2006diagnosis,qi2013fault,choi2018detecting}. The effectiveness of this method has already been demonstrated in our previous studies \cite{bluebox}. Such a preliminary result shows the robustness of this method under the bounded modeling uncertainties and disturbances. However, \cite{bluebox} focused on ensuring the security of the vehicle controller using a redundant approach. In this work, the FDI focuses on identifying and responding to sensor physical faults.

The typical SVSF estimator consists of two stages: prediction and correction. In one iteration, a model-based prediction function $\hat{\mathcal{F}}$ generates a priori state estimation $\hat{\bm{x}}_{k|k-1}$ first. A discrete corrective action is then taken by adding a corrective gain $\bm{K_k}$. The corrective gain is used to both guarantee the stability of the estimator and also rectify the bounded estimation error. Subsequently, the updated posteriori estimation $\hat{\bm{x}}_{k|k}$ and state measurement $\bm{z}_{k|k}$ carry out the next iteration. In order to implement SVSF properly, we rewrite the nonlinear MAV model to a discrete form \cite{bluebox}
\begin{equation}
\begin{aligned}
	\bm{x}_{k+1} &= \mathcal{F}(\bm{x}_k,\bm{u_k},\bm{\tilde{d}_k}) ,\\
	\bm{z}_{k+1} &= \bm{H}\bm{x}_k + \bm{\tilde{n}_k},
\end{aligned}
\end{equation}
where the vehicle states are given by $\bm{x} = [\bm{P}, \bm{V}, \bm{\Theta}, \bm{\omega}^b]^T$, wherein $\bm{\Theta} = [\phi,\theta,\psi]^T$ represents vehicle's 3-axis Eular angle; 
$k$ is the time step; 
the vector $\bm{u} = [u_1, u_2, u_3, u_4]^T$ denote the system input which is the input voltages of the four motors; 
$\bm{\tilde{d}}$ is the external disturbance; 
$\bm{z}$ is the state measurement; 
$\bm{H} = \textbf{{I}}_{12}$ where $\textbf{{I}}_{12}$ is a twelfth-order identity matrix; 
$\bm{\tilde{n}}$ is the measurement noise.

Following the above discrete model, a prediction function $\hat{\mathcal{F}}$ can be defined accordingly. 


With $\hat{\mathcal{F}}$, the prediction stage can be expressed by
\begin{equation}
\begin{aligned}
	\hat{\bm{x}}_{k|k-1} &= \hat{\mathcal{F}}(\hat{\bm{x}}_{k-1|k-1},\bm{u}_{k-1}),\\
	\bm{e}_{k|k-1} &= \bm{z}_k - \hat{\bm{H}} \hat{\bm{x}}_{k|k-1}.
\end{aligned}
\end{equation}
where
$\bm{\hat{\bullet}}$ denotes the corresponding estimation variable. 
The estimation result is defined by $\hat{\bm{z}}_{k} = \hat{\bm{H}} \hat{\bm{x}}_{k}$.

The correction stage is given by
\begin{equation} \label{eq:SVSF correction}
\begin{aligned}
	&\bm{K}_k =  \hat{\bm{H}}_k^{-1} ( |\bm{e}_{k|k-1}| + \gamma |\bm{e}_{k-1|k-1}| ) \circ sgn(\bm{e}_{k|k-1}), \\
	&\hat{\bm{x}}_{k|k} = \hat{\bm{x}}_{k|k-1} + \bm{K}_k,\\
	&\bm{e}_{k|k} = \bm{z}_k - \hat{\bm{H}}_k \hat{\bm{x}}_{k|k}.
\end{aligned}
\end{equation}

The stability proof of the estimator is similar to that in \cite{habibi2007smooth} with the different state variables. 

When the vehicle is subject to sensor failure, the incorrect feedback can induce discrepancies between estimation result and actual sensor readings, generating non-negligible residual variations: 
\begin{equation}
\bm{r} = \sum_{i=k}^{k+N} (\bm{z}_i - \hat{\bm{z}}_{i})(\bm{z}_i - \hat{\bm{z}}_{i})^T,
\end{equation}
where $\bm{r}$ is the residual vector dedicated to each state; $1/N$ is the updating frequency.

For FDI, if the residual rises beyond a certain threshold, a fault/attack is presumably present and will be reported, consequently starting the execution of isolation logic. Such thresholds can be determined by the system physical bandwidth and control input constraints. Experimentally acquisition of sensor readings under the critical stable flight also can guide the threshold selection.

\subsection{Fine-grained Sensor FDI} \label{subsec: Novel}
Detecting abnormalities in the system states relies solely on the interpretation of the residual, which can expose the presence of the fault/attack but cannot explicitly pinpoint its specific source in hardware. This is mainly because some of the state parameters cannot be measured directly, such as attitude information, i.e., roll, pitch, and yaw angles. Instead, they are usually updated through a sensor fusion algorithm that combines the feedback from multiple sensors in a certain sampling frequency. This synthesis of sensor feedback is moderately beneficial in that it takes advantages from each individual sensor's unique characteristics. Such a complementary effect reduce sensor error and noise significantly. However, it also weakens the impact of any single sensor and mitigates the impact of the sensing faults to some degree due to the coupled vehicle dynamics. This general issue limits the efficiency and accuracy of typical model-based FDI methods as they highly rely on sensor fusion results, which motivates us to improve the current design. In this subsection, we will propose to use the variations of physical parameters of the vehicle (mass/inertia) to address this problem.

Since the physical parameters of the vehicle are usually not changed significantly during the real flight, if a sensor fault/attack appears, we can track the unexpectedly varying model parameters in reverse and trace the source of the variations to the specific faulty components. For our estimator, due to the zero-width smoothing layer in $\mathbf{K}_k$, the modeling error is referred to as a chattering signal.

The a priori chattering of $\mathbf {Ch}=[\text{CH}_{\dot{z}}, \text{CH}_p, \text{CH}_q, \text{CH}_r]^T$ is defined as
\begin{equation}
\begin{gathered}
	\mathbf {Ch}_{{k|k-1}} = 
	dt
	\begin{bmatrix}

		\Delta_{(\frac{C_L}{m})}(C\psi_{k-1}C\phi_{k-1}) f_z\\

		\Delta_{(J_1)}q_{k-1}r_{k-1}+\Delta_{(\frac{C_L}{J_x})}\tau_x\\
		\Delta_{(J_2)}p_{k-1}r_{k-1}+\Delta_{(\frac{C_L}{J_y})}\tau_y\\\
		\Delta_{(J_3)}p_{k-1}q_{k-1}+\Delta_{(\frac{C_D}{J_z})}\tau_z\
	\end{bmatrix},
\end{gathered}
\end{equation}
where $C\bullet$ and $S\bullet$ denote $cos(\bullet)$ and $sin(\bullet)$ respectively for compact notation; 
$J_x, J_y, J_z, m$ are the moment of inertias and mass of the vehicle; 
$J_1 = \frac{J_y-J_z}{J_x}$, $J_2 = \frac{J_x-J_z}{J_y}$, $J_3 = \frac{J_x-J_y}{J_z}$. 

\begin{figure*}[!h]
\begin{center}
\includegraphics[trim = 0mm 0mm 0mm 0mm, clip,width=1.9\columnwidth]{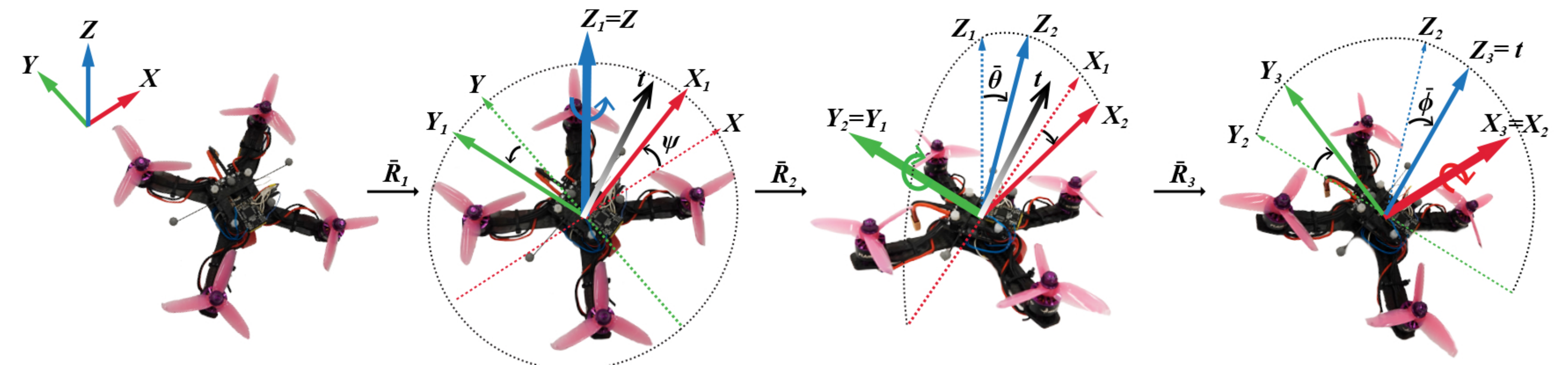}
\caption{Geometric principles of attitude feedback recover when the inertial sensor has been isolated. The rotations along Z-Y-X axes can be derived using the changes in the relative Cartesian axis vectors respectively as shown.}
\label{fig:Recovery_3rot}
\end{center}
\vspace{-0.2in}
\end{figure*}

From the above formulation, such a priori chattering is caused by the magnitude of the model uncertainties and can be used to point out the source and magnitude of modeling error \cite{habibi2007smooth}. In contrary, if the vehicle model is trusted, it infers related sensing/actuation flaws accordingly. Therefore, it can be utilized to pinpoint the faulty components of the vehicle. 
In normal flights, the chattering is bounded and can be used to refined model parameters through a short time on-line calibration. The calibrated model contains a set of meaningful physical parameters relevant to the inception of the IMU working status in the MAV system. Specifically, the IMU fault/attack causes additional forces and torques output, yielding a substantial transient error in the estimation of the corresponding physical parameters. This error can then be employed to pinpoint the compromised IMU sensor as its measurements are directly related to these parameters. On the other hand, for underactuated MAVs like quadrotors, because the position measurements are largely independent of the physical parameters being estimated, position sensor spoofing will only cause $\bm{r}$ and corresponding $\bm{e}$ terms to change, showing minor effects on $\mathbf {Ch}$. The effectiveness of using chattering signal for IMU failure identification is demonstrated in section.\ref{sec:result}.

\section{Flight Recovery} \label{sec:FR}

The vulnerabilities of MAVs' onboard sensors are different: inertial components are sensitive to vibrations; magnetic components cannot be resistant to magnetic field distortion; position sensing is usually derived from external references (satellites, landmarks or cameras), relying on communication quality. Because of these differences, for adversaries, it is almost impossible to develop a coordinated attack that covers all sensors \cite{guo2019position}. This fact gave us the confidence to use the trusted sensors that have not been isolated to complement and accommodate the faults of the others. This complementary approach does not require hardware redundancy, which fits those small sized aerial vehicles that have SWaP constraints. 

In this section, we present our complementary method for IMU fault/attack recovery. In particular, due to the geometric correlations of the vehicle dynamics, position feedback can be used to provide an alternative attitude approximation for flight control. During the recovery process, following the fine-grained FDI, the compromised IMU readings will be totally rejected. The recovery logic instead uses the rest of unattacked sensors for attitude estimation. The successful recovery shown in section.\ref{sec:result} validates the performance of the proposed FR method.

For MAVs, if the IMU faults are detected and isolated, besides the absolute yaw angle measured by compass, most of the body frame information becomes unreachable. It is a severe fault as the vehicle is certain to lose attitude stability. To ensure flight safety, our geometric approach associates attitude with such limited remaining feedback, i.e., $x,y,z,\psi$ to approximate the current attitude and allow control.

Inspired by the MAV geometric control \cite{lee2010geometric}, the essence of MAV attitude generation is the direction change of the thrust vector. 
Since the thrust vector is always perpendicular to the rotor-disk plane on rotary-winged vehicles, it determines the roll and pitch attitude specifically. Based on this phenomenon, we can use the following geometrical approach for attitude recovery. 

From equation.(\ref{eq:body_dynamics}), in the inertial frame $\bm{T} = m \ddot{\bm{P}} - m \bm{g}$. It is normalized as $\bm{t} = \dfrac{\bm{T}}{|\bm{T}|} = [t_1,t_2,t_3]^T$ for later calculation. We ignored the small body damping here for our test platform. The heading direction of the vehicle defines the first coordinate transformation $\bm{\bar{R}_1}$ as shown in Fig.\ref{fig:Recovery_3rot}, 
\begin{equation}
	\bm{\bar{R}_1} = \begin{bmatrix}
		C \psi 		& S \psi 		& 0 \\
		-S \psi 	& C \psi 		& 0 \\
		0 			& 0 			& 1 \\
	\end{bmatrix},
\end{equation}
where $\psi \in (0, 2\pi)$.

Through $\bm{\bar{R}_1}$ rotation, the original north-east-up coordinate $X Y Z$ is transferred to an auxiliary coordinate $X_1  Y_1  Z_1$. Then, the vehicle tilting angles along the pitch and roll direction are determined by $\bm{\bar{R}_2}$ and $\bm{\bar{R}_3}$, respectively:
\begin{equation} \label{eq: R2_bar and R3_bar}
\begin{aligned}
	\bm{\bar{R}_2} &= \begin{bmatrix}
		C \bar{\theta} & 0 & S \bar{\theta} \\
		0 & 1 & 0 \\
		-S \bar{\theta} & 0 & C \bar{\theta}\\
	\end{bmatrix} \\ 
	&= \begin{bmatrix}
		\bm{Z_1 \cdot Z_2} & 0 & \bm{Y_1} \cdot (\bm{Z_1 \times Z_2}) \\
		0 & 1 & 0 \\
		-\bm{Y_1} \cdot (\bm{Z_1 \times Z_2}) & 0 & \bm{Z_1 \cdot Z_2} \\
	\end{bmatrix}, \\
	\bm{\bar{R}_3} &= \begin{bmatrix}
		1 & 0 & 0 \\
		0 & C \bar{\phi} & -S \bar{\phi} \\
		0 & S \bar{\phi} & C \bar{\phi} \\
	\end{bmatrix} \\ 
	&= \begin{bmatrix}
		1 & 0 & 0 \\
		0 & \bm{Z_2 \cdot t} & -\bm{X_2} \cdot (\bm{Z_2 \times t}) \\
		0 & \bm{X_2} \cdot (\bm{Z_2 \times t}) & \bm{Z_2 \cdot t} \\
	\end{bmatrix}.
\end{aligned}
\end{equation}

Through rotation $\bm{\bar{R}_2}$ and $\bm{\bar{R}_3}$, the coordinate transformation follows the sequence of $X_1 Y_1 Z_1 \rightarrow X_2  Y_2  Z_2 \rightarrow X_3 Y_3 Z_3$, wherein $X_3  Y_3  Z_3$ is the current body frame coordinate. 
The corresponding $\bar{\theta}$ and $\bar{\phi}$ in $\bm{\bar{R}_2}$ and $\bm{\bar{R}_3}$ provide attitude feedback which is generated based on position and heading feedback in the inertial frame. In normal conditions, both $\bar{\theta},\bar{\phi} \in (-\dfrac{\pi}{2}, \dfrac{\pi}{2})$. 

With $\psi$ and $\bm{t}$, we know $X_1 = [C\psi, S\psi, 0]^T$ and $\bm{t} = [t_1,t_2,t_3]^T$. 
The unknown terms in $\bm{\bar{R}_2}$ and $\bm{\bar{R}_3}$ matrices are

\begin{equation} \label{eq:pseudo roll_pitch}
\left\{
\begin{aligned}
C \bar{\theta}  &= \bm{Z_1 \cdot Z_2} = \dfrac{\bm{t \times Y_1}}{|\bm{t \times Y_1|}} = \dfrac{t_3}{\lambda(\bm{t},\psi)} \\
S \bar{\theta}  &= \bm{Y_1} \cdot (\bm{Z_1 \times Z_2}) = \dfrac{\bm{t \times X_1}}{|\bm{Y_1 \times t|}} = \dfrac{t_1 C\psi + t_2 S\psi}{\lambda(\bm{t},\psi)} \\
C \bar{\phi} 	  &=	\bm{Z_2 \cdot t} = \lambda(\bm{t},\psi) \\
S \bar{\phi}	  &= -\bm{X_2} \cdot (\bm{Z_2 \times t}) = \bm{Y_2 \cdot t} = -t_1 S\psi + t_2 C\psi \\
\end{aligned}
\right. ,
\end{equation}
where $\lambda(\bm{t},\psi) = \sqrt{t_3^2 + (t_1 C\psi + t_2 S\psi)^2}$. 

From equation.(\ref{eq:pseudo roll_pitch}), the complimentary attitude feedback $\bar{\phi},\bar{\theta}$ can be calculated mathematically using inverse trigonometric functions $arccos(\bullet)$ and $arcsin(\bullet)$. 

By taking advantage of the fact that the pose of the vehicle determines the thrust orientation, the proposed method above is able to recover attitude control with limited sensor readings. 
The proposed method can also be generalized to any sensor that can provide the thrust vector measurement, such as visual, acoustic, and laser sensors. For implementation, if the frequency of the recovered feedback is insufficient for control, gain tuning or digital control is required.
\section{Simulation and Experimental Results} 
\label{sec:result}
For the purpose of validation, we deployed our method onto a customized simulation tool \cite{fei2019flappy}, which was used to solve the nonlinear dynamics of the vehicle. The physical parameters of the vehicle used in the simulation were obtained by conducting system identification on the test platform. Each component of such platform has been measured, weighted, and modeled in CAD software, and mass/inertia parameters are calculated accordingly. In the simulation, we verified the proposed FDI and FR. Then, we experimentally demonstrated run-time attitude recovery with a compromised IMU, which is still considered challenging for most of the current fault-tolerant schemes for MAVs \cite{blanke2006diagnosis,qi2013fault,avram2015imu,qi2007application, avram2017quadrotor,choi2018detecting}. The transmission delay will affect the FR performance. In our setup, it is about 0.025s which slows down the effective control frequency and causes control error. 

\subsection{Simulation Validations} 

\subsubsection{FDI Validation} 
\begin{figure}[b]
\begin{center}
\includegraphics[trim = 0mm 0mm 0mm 0mm, clip,width=0.95\columnwidth]{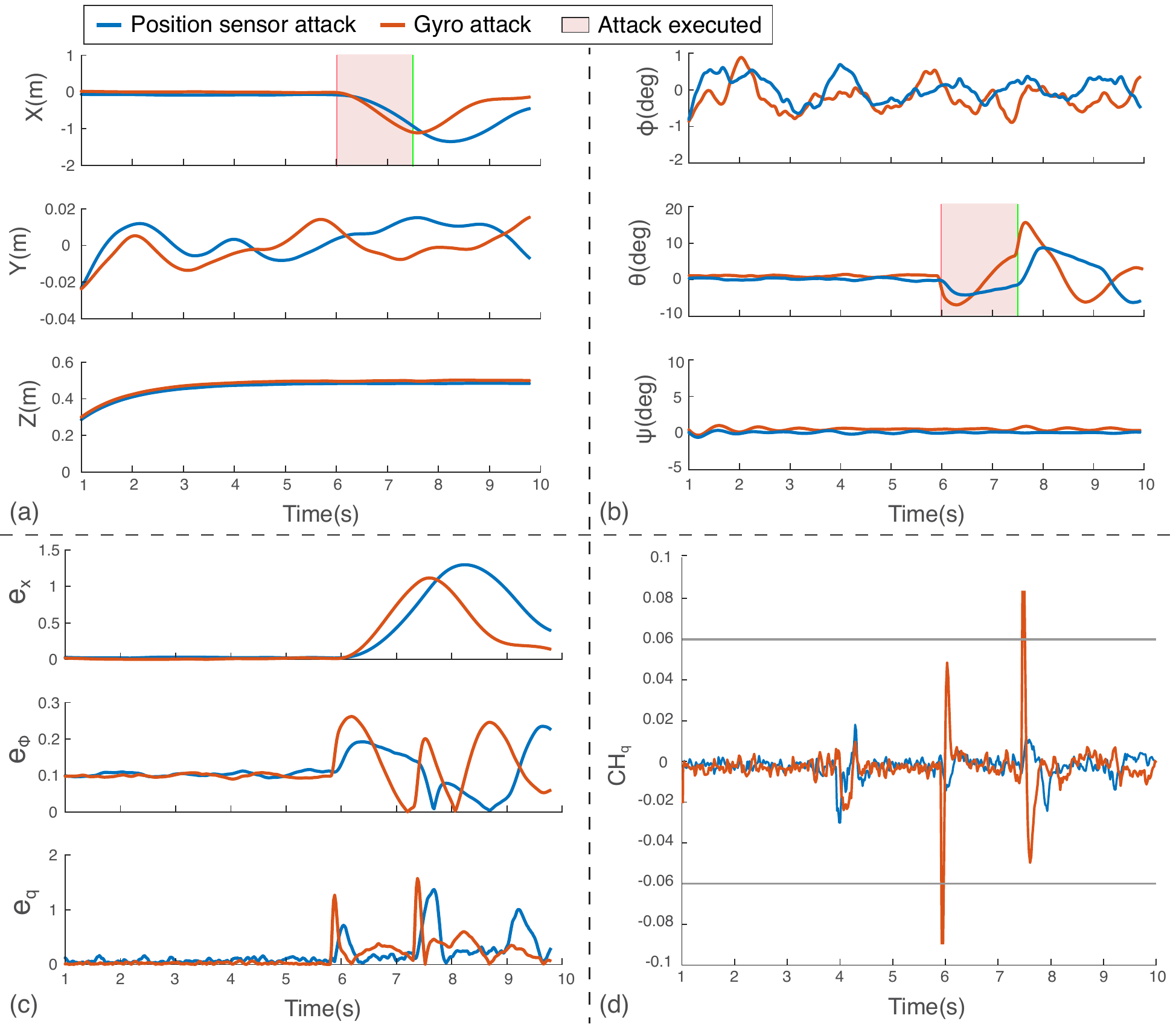}
\caption{Feasibility of the proposed FDI design under different sensor attacks is shown in this figure. To demonstrate, we added additive bias onto VICON and gyroscope readings which are tuned to achieve similar flight patterns (position and attitude) as shown in (a) and (b). The corresponding state estimation errors that dominate the residual and chattering information are shown in (c) and (d), respectively. 
}
\label{fig:isolation_sim_result}
\end{center}
\vspace{-0.2in}
\end{figure}
In this test, the VICON and IMU readings were attacked individually to show the effectiveness of the proposed FDI design in pinpointing the compromised sensor under similar abnormal MAV behaviors. In particular, the x-position feedback and the corresponding axis' gyroscope reading were attacked. For flight control, a position-attitude cascade PD controller was used. As shown in Fig.\ref{fig:isolation_sim_result} (a-b), these two different types of sensor attack generated close flight paths and attitude responses. Such a scenario provides a challenge to differentiate those attacks. For FDI, we first use SVSF as a traditional state estimator to generate residual of the related vehicle states. Such a residual can detect anomalies. However, it cannot pinpoint the exact faulty sensor as the state estimation errors of these two scenarios are also pretty close as shown in Fig.\ref{fig:isolation_sim_result} (c), which yields indistinguishable residual accordingly. As an underactuated system, such MAV's coupled dynamics induces the multiple flight states to change from tampering with single feedback. Thus, just  checking residual usually cannot explicitly determine the compromised parts, potentially causing incorrect isolation. To improve the FDI accuracy, we propose using the chattering signals $\text{CH}_q$ to provide an effective way to identify the specific compromised sensor. 

Chattering signals, rendered by physical model variations, offer a clear solution to address this problem since the vehicles' physical parameters (mass and inertia) usually do not change significantly during the flight. As shown in Fig.\ref{fig:isolation_sim_result} (d), because tampering with gyroscope measurements cause the body inertia estimation to diverge while position spoofing does not, the chattering signal appears more sensitive when the IMU is faulty than that the position sensor is. Consequently, gyroscope attacks can be identified explicitly and reliably by the significant fluctuations in chattering signal, differentiating from position sensor attack scenario. Here, $\pm 0.06$ (grey lines) is chosen as the threshold to detect gyroscope abnormalities.

\begin{figure}[b]
\begin{center}
\includegraphics[trim = 0mm 0mm 0mm 0mm, clip,width=0.95\columnwidth]{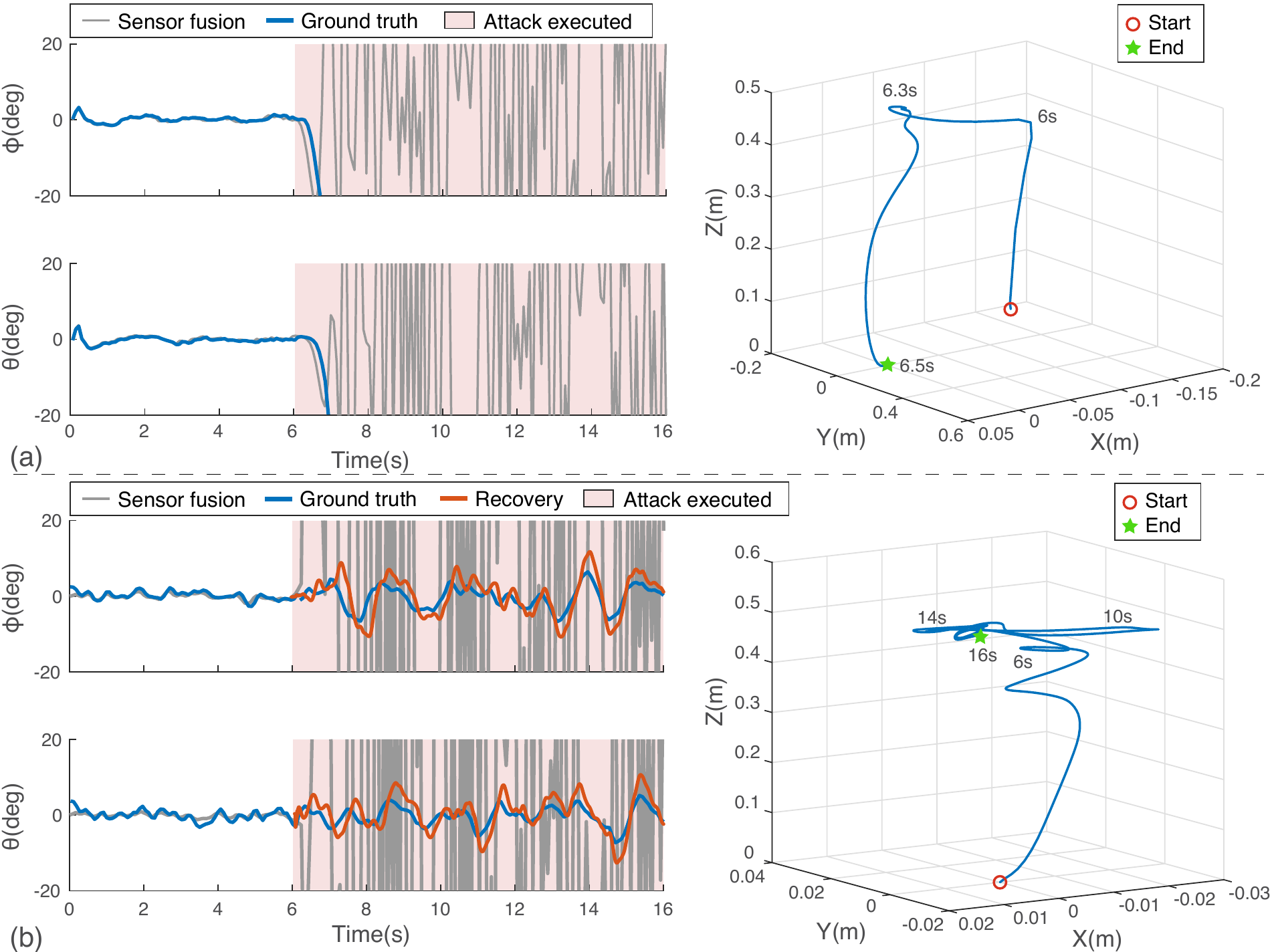}
\caption{(a) shows the effects of IMU feedback being fully isolated in hover flight without FR function: as the attack triggered, the attitude feedback is totally diverging. Without such available feedback, the vehicle lost its flight stability immediately. In test (b), the proposed FR algorithm has turned on. With FR, the vehicle can maintain its stability without IMU. Such attacks induced oscillated IMU readings, which causes inaccurate sensor fusion result. Ground truth is from VICON records with an inevitable delay. Due to the relatively low update frequency of the restored attitude feedback, position and attitude control shows errors but remains in a certain bound. 
}
\label{fig:recovery_sim_result}
\end{center}
\vspace{-0.2in}
\end{figure}



\begin{figure*}[!t]
\begin{center}
\includegraphics[trim = 0mm 0mm 0mm 0mm, clip,width=1.95\columnwidth]{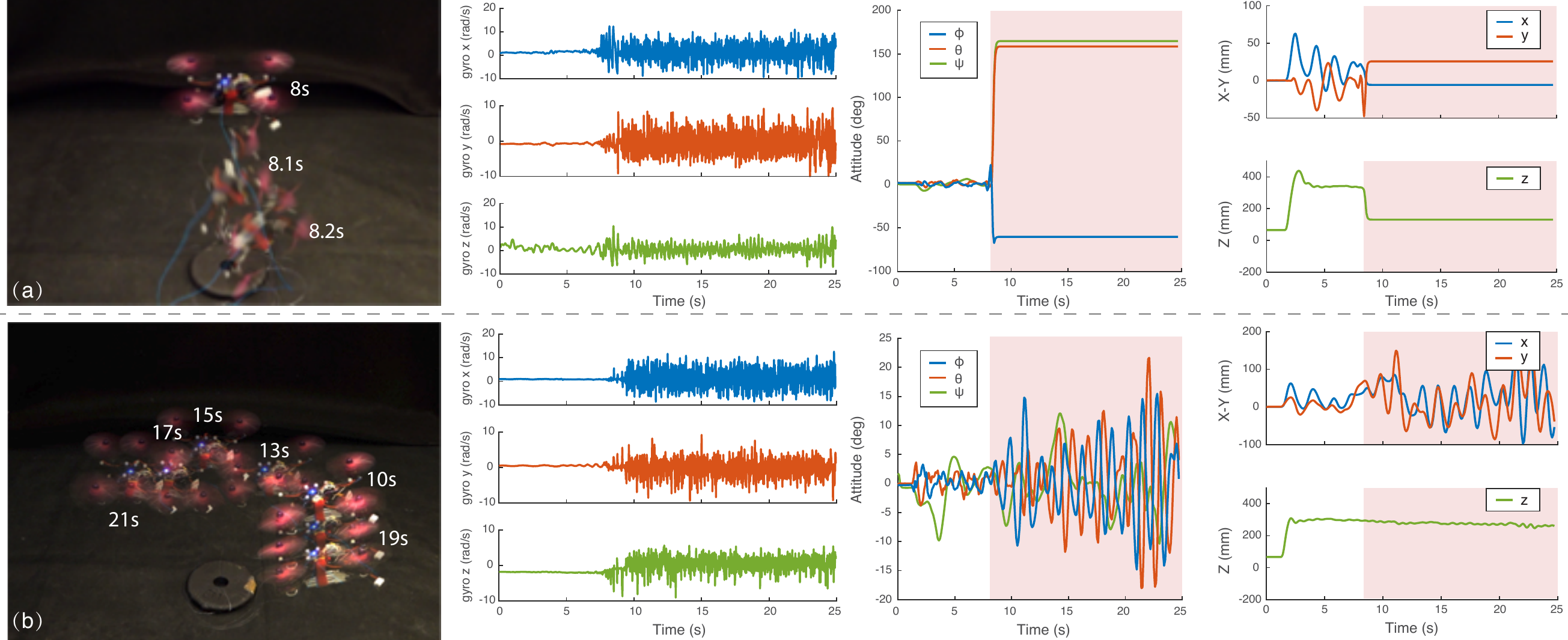}
\caption{(a) shows the IMU attack without FR. From left to right, the composed time sequential result, the tricked IMU readings like \cite{son2015rocking}, the attitude response, and flight trajectory are shown respectively. These severely oscillated IMU signals affect attitude stabilization significantly, the vehicle crashing immediately about 0.2 seconds. Here, only gyro's raw data is plotted for the demonstration of IMU attack, though the not shown accelerometer measurements were compromised simultaneously. (b) shows the IMU attack with FR. With FR, since the compromised IMU readings has been totally rejected, the attitude stability of the vehicle has ensured by the complementary feedback accordingly. The pink area indicates that FDI detected an IMU attack and the corresponding sensor readings had been isolated. 
}
\label{fig:exp_result1_sensorfault}
\end{center}
\vspace{-0.2in}
\end{figure*}

\subsubsection{FR Validation} 
This test assesses the effectiveness of the proposed method of attitude feedback recovery and re-stabilization of the vehicle. The prerequisite of the execution of FR is that the proposed FDI module detects the corresponding IMU faults. The control law is the same as the last test. When an IMU attack is generated, the default attitude sensing (sensor fusion result) is totally divergent. FDI detected anomalies and isolated IMU measurements accordingly such that feedback from the IMU (particularly, measurements of accelerometer and gyroscope) is completely rejected and the vehicle drops immediately without any replacement feedback as shown in Fig.\ref{fig:recovery_sim_result}.(a). In comparison, utilizing the recovery logic with the remaining compass sensor for attitude estimation, the vehicle was able to maintain its attitude stabilization and stay within a bounded area (a 0.2m circular around area for 16 seconds in this test). While the position tracking of the vehicle is not ideal due to the lower feedback frequency inherent with this method, it achieves safe flight as demonstrated in Fig.\ref{fig:recovery_sim_result}.(b).  

\subsection{Experimental Results and Discussion} 
This experiment aims to counter the severe oscillations in IMU measurements, which is a significantly threatening attack for MAVs as presented in \cite{son2015rocking} since traditional fault-tolerant strategies become ineffective under such attacks though they have shown high reliability in many other tasks \cite{blanke2006diagnosis,qi2013fault,choi2018detecting,bluebox}. A PD control law is adopted for position tracking and attitude stabilization as it was used in the simulation. For demonstration, we replicate the sensor failure pattern presented in \cite{son2015rocking} by maliciously aliasing a high-low-frequency signal on the IMU sensor via trojan input to mimic its resonating output as shown in Fig.\ref{fig:exp_result1_sensorfault}. Under such attack, the compromised IMU needs to be isolated immediately. Then, the position information could be used to provide alternative attitude feedback to stabilize the vehicle. During the experiment, we first triggered the fault without a fault-tolerant scheme: the vehicle lost stability immediately as shown in Fig.\ref{fig:exp_result1_sensorfault}. In contrast, utilizing the proposed FDI and FR, the MAV demonstrated stable hovering flight with the compromised IMU. From the experiment results shown in Fig.\ref{fig:exp_result1_sensorfault}, the vehicle can maintain its stability for sustained flight. Compared to simulated flight, the onboard implementation required less precise lightweight sensors and processors leading to slightly larger numerical error and subsequent position drift in the experimental test, but it still provided enough time for a safe landing or other safe reactions. It is worth noting that the position information used to make the necessary attitude feedback estimates was updated with higher accuracy and frequency than common GPS systems provide. To date, GPS systems with update frequencies of 50 Hz and above are commercially available. 

\section{Conclusion} \label{sec: Conclusion}
In this work, a fault-tolerant design is proposed for MAVs against IMU compromised scenarios. The proposed design included two main parts: FDI and FR. FDI was performed, which took into account the vehicle states and modeling uncertainties to accurately determine the faulty sensors and execute isolation of compromised feedback. Afterward, for flight recovery with incomplete feedback, we proposed a geometrical method that relies on the remaining position and heading information to estimate attitude feedback needed to support flight control, without control program modification. The proposed MAV sensor fault-tolerant design has been validated through both simulation and experimental tests using a small-sized quadcopter platform. The results show that this novel design can successfully detect and respond to compromised IMUs, allowing the vehicle to recover stabilized flight without IMU readings. Though there is some loss of position control, i.e., the position drifts over time, the more critical attitude control is maintained allowing the vehicle to avoid an otherwise imminent crash. The future work following this effort could be porting this fault-tolerant design to other different types of MAVs, e.g., fixed-wing or flapping-wing MAVs, and incorporating it with other fault-tolerant methodologies.

\balance
\bibliography{FDI}
\bibliographystyle{IEEEtran}

\balance
\end{document}